%% file: acl_latex.tex
\pdfoutput=1
\documentclass[11pt]{article}

\usepackage[final]{acl}

\usepackage{times}
\usepackage{latexsym}
\usepackage{booktabs}
\usepackage{graphicx}
\usepackage{subfigure}
\usepackage{amsmath}
\usepackage{amssymb}
\usepackage{mathtools}
\usepackage{amsthm}
\usepackage{adjustbox}
\usepackage{multirow}

\usepackage{tcolorbox}
\tcbuselibrary{listings,breakable,skins}

\newcommand{\method}[0]{\textsc{DoubleDipper}}

\usepackage[T1]{fontenc}
\usepackage[utf8]{inputenc}

\usepackage{microtype}

\usepackage{inconsolata}

\usepackage{graphicx}

\title{\method{}: Improving Long-Context LLMs via Context Recycling}

\author{Arie Cattan$^{1,2}$\thanks{Work done during an internship at Google.} \quad 
Alon Jacovi$^{2}$ \quad 
Alex Fabrikant$^{3}$ \quad
Jonathan Herzig$^{2}$ \quad \\ 
\textbf{Roee Aharoni}$^{2}$ \quad 
\textbf{Hannah Rashkin}$^{3}$ \quad
\textbf{Dror Marcus}$^{2}$ \quad  \\
\textbf{Avinatan Hassidim}$^{2}$ \quad 
\textbf{Yossi Matias}$^{2}$ \quad 
\textbf{Idan Szpektor}$^{2}$\quad 
\textbf{Avi Caciularu}$^{2}$ \\
$^{1}$Bar-Ilan University \quad $^{2}$Google Research \quad $^{3}$Google DeepMind \\  
\texttt{cattana@google.com} \\ 
}

\begin{document}
\maketitle

\input{sections/00_abstract}
\input{sections/01_intro}

\input{sections/02_bg}

\input{sections/03_method}

\input{sections/04_experiments}

\input{sections/05_results}

\input{sections/06_analysis}

\input{sections/07_conclusion}

\input{sections/08_limitations}

\bibliography{anthology,custom}

\appendix

\section{Appendix}

\section{Prompts}
\label{app:prompts}

Figure~\ref{fig:qg_prompt} shows the zero-shot prompt we use for generating the question-answer pairs in \method{}. 
For the QA prompts, we use the same instructions and prompt template as the original papers (Lost-in-the-middle and FLenQA) and add a simple line for the instructions in other multi-hop QA datasets: \emph{``Please answer the question based on the given passages below.''}. For MuSique, since the dataset includes  questions that are not answerable, we add the following sentence to the prompt: \emph{``If the question can't be answered given the given passages, please write "unanswerable"''}.

\input{prompts/qg_prompt}

% \section{Identification of Supporting Passages}
% \label{app:retrieval}

% Table~\ref{tab:ret} presents the F1 results of the tested models for the supporting relevant passages identification. Without any demonstration (zero-shot), all open source models achieve a poor performance, while \method{} significantly improves performance (+37.8 F1 for Llama2 70B, +38 F1 for Gemma 7B, +30.5 F1 for Mistral). 

% \input{tables/retrieval}

\section{Additional Results}
\label{app:additional_results}

Table~\ref{tab:appendix_results} shows the QA performance of the  baseline, \emph{Zero-shot  +  Evidence  Retrieval} and \method{} on our evaluation set when applied to Gemma 2B, Gemini Nano and GPT-4o-mini. Her

\input{tables/appendix_results_dd_vs_baseline}

\section{Lost-in-the-middle}
\label{app:lost}

\begin{figure*}[t]
    \centering
    \includegraphics[width=.98\textwidth]{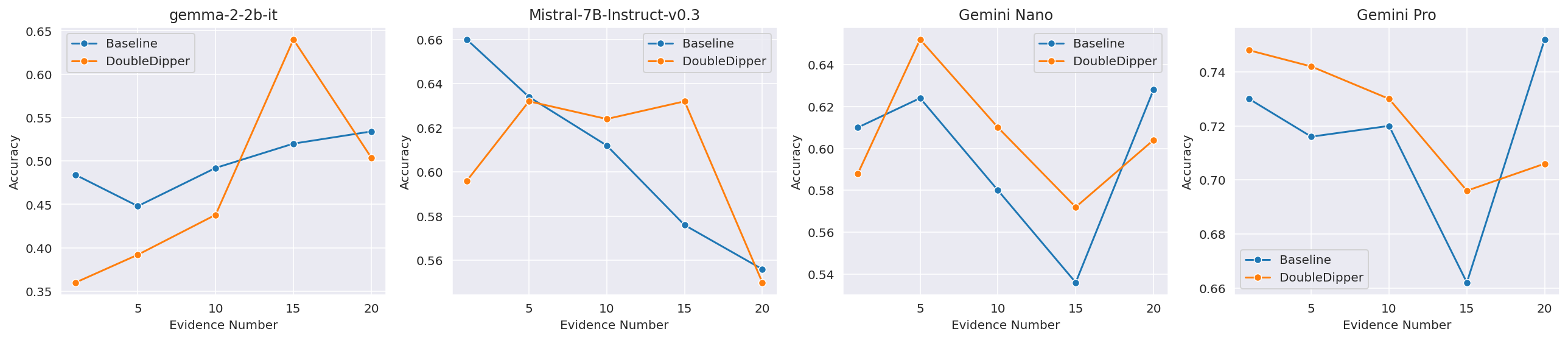}
    \caption{Performance (accuracy) of Gemma 2 2B, Mistral 7B, Gemini Nano and Gemini Pro with and without \method{} on our sample of the Lost-in-the-middle dataset~\citep{Liu2023LostIT} according to the position of the document that contains the answer.}
    \label{fig:lost_app}
\end{figure*}

Figure~\ref{fig:lost_app} shows the QA accuracy of the models Gemma 2 2B, Mistral 7B, Gemini Nano and Gemini Pro on our subset of the ``Lost-in-the-middle'' dataset.

\section{Analysis}
\label{app:analysis}

\subsection{Impact of the Number of Demonstrations in \method{}}
\label{app:subsec:k}

\input{tables/few_shots_detailed}
Table~\ref{tab:results_k_detailed} presents the results of \method{} with 1, 3 (main experiment in the paper), 5 and 10 generated demonstrations. For all these experiments, the demonstrations were generated by Gemma 2 2B.

\begin{figure*}[t]
    \centering
    \includegraphics[width=.98\textwidth]{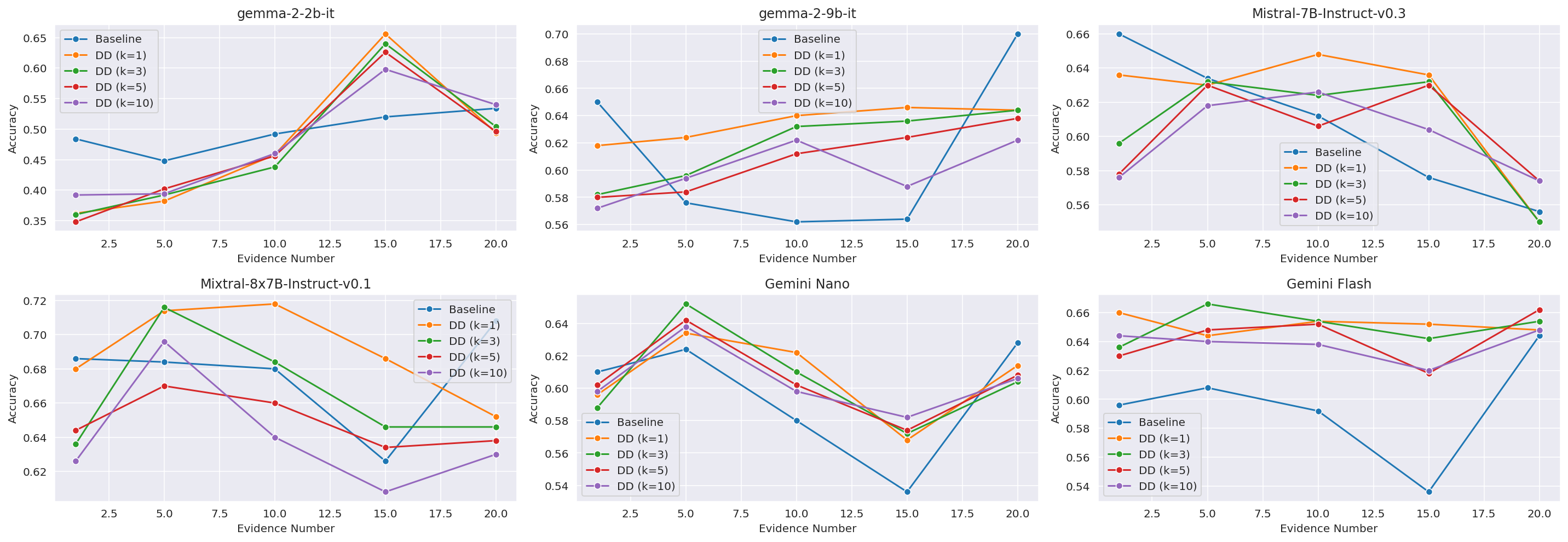}
    \caption{Performance (accuracy) of several models with \method{} according to the number of self-generated demonstrations in the prompt. }
    \label{fig:lost_per_k}
\end{figure*}

Figure~\ref{fig:lost_per_k} shows the QA accuracy of \method{} on ``Lost'' according to the position of the relevant passage for each $k \in \{1, 3, 5, 10\}$. 

\subsection{Impact of the few-shot generator}
\label{app:subsec:few_shot_model}

Table~\ref{tab:few_shot_generator_detailed} presents the detailed QA performance of all models with different models for generating \method{}'s demonstrations. 
As mentioned in the paper~(Section~\ref{sec:analysis}), generating the demonstrations with the best model (ie. Gemini Pro) achieves the best performance overall. 

\input{tables/dd_few_shot_generator_detailed}

\paragraph{Can we generate demonstrations with smaller LLMs?} 
We next study whether smaller, more efficient models could be used to generate demonstrations for \method{}. To do this, we created demonstrations using two smaller LLMs, Gemma 3 1B~\citep{Kamath2025Gemma3T} and Qwen 2.5 0.5B~\citep{Bai2023QwenTR}, and provid them to a subset of our evaluation models (Llama 3.1, Gemma 2 9B, Gemma 2 27B, and Mistral Nemo). The results show that \method{} maintains its advantage over baselines, exhibiting only a modest performance drop of less than 2 points compared to using demonstrations from the larger Gemma 2 2B. This finding suggests that \method{} remains effective even when its demonstration generation component is replaced with more lightweight models.

\paragraph{Can \method{} benefit from incorrect demonstrations?}

To answer this question, we prompt Gemma 2 2B to generate a question and an \textit{incorrect} answer and provide these demonstrations to a sample of our tested models (Llama 3.1, Gemma 2 9B, Gemma 2 27B and Mistral Nemo). 
The results are mixed and not conclusive: some models barely benefit or suffer from incorrect demonstrations (+1.6 for Gemma 2 9B, -0.4 for Gemma 2 27B), while others somehow benefit from incorrect demonstrations (+4.8 for Llama 3.1 8B and +5.1 for Mistral Nemo over the baseline). Critically, however, all models still perform substantially worse than \method{} with correct demonstrations. A possible explanation for the unexpected gains is that even incorrect examples provide useful structural guidance for the task format, a phenomenon observed in prior work~\citep{min-etal-2022-rethinking}.

\subsection{Impact of the identification of supporting paragraphs in the QA generation}
\label{app:impact_retrieval}
Table~\ref{tab:dd_without_evidence} compares the performance of \method{} to \method{} without evidence identification.

\input{tables/dd_without_evidence}

\subsection{In-Context-Learning}
\label{app:icl_with_squad}

Table~\ref{tab:appendix_icl} presents the results of our tested models when prepended with three in-context demonstrations, taken from the Squad 2.0 dataset~\citep{rajpurkar-etal-2018-know}.

\input{tables/appendix_icl_squad_results}

\end{document}

%% file: sections/00_abstract.tex
\begin{abstract}

Despite recent advancements in Large Language Models (LLMs), their performance on tasks involving long contexts remains sub-optimal. 
In this work, we propose \method{}, a novel In-Context-Learning method that automatically generates few-shot examples for long context QA tasks by \textit{recycling} contexts. Specifically, given a long input context (1-3k tokens) and a query, we generate additional query-output pairs from the given context as few-shot examples, while introducing the context only once. This ensures that the demonstrations are leveraging the same context as the target query while only adding a small number of tokens to the prompt. We further enhance each demonstration by instructing the model to \textit{explicitly} identify the relevant paragraphs before the answer, which improves performance while providing fine-grained attribution to the answer source.  
We apply our method on multiple LLMs and obtain substantial improvements (+16 absolute points on average across models) on various QA datasets with long context. Surprisingly, despite introducing only single-hop ICL examples, LLMs successfully generalize to multi-hop long-context QA using our approach.

\end{abstract}

%% file: sections/01_intro.tex
\section{Introduction}
\label{sec:intro}

Long contexts are prevalent in various domains, ranging from legal documents and scientific articles to lengthy reports and novels. These may consist of a single extensive document or multiple passages, typically retrieved through specific retrieval mechanisms (e.g., RAG \cite{rag2020}).

\begin{figure}[t]
    \centering
    \includegraphics[width=.48\textwidth]{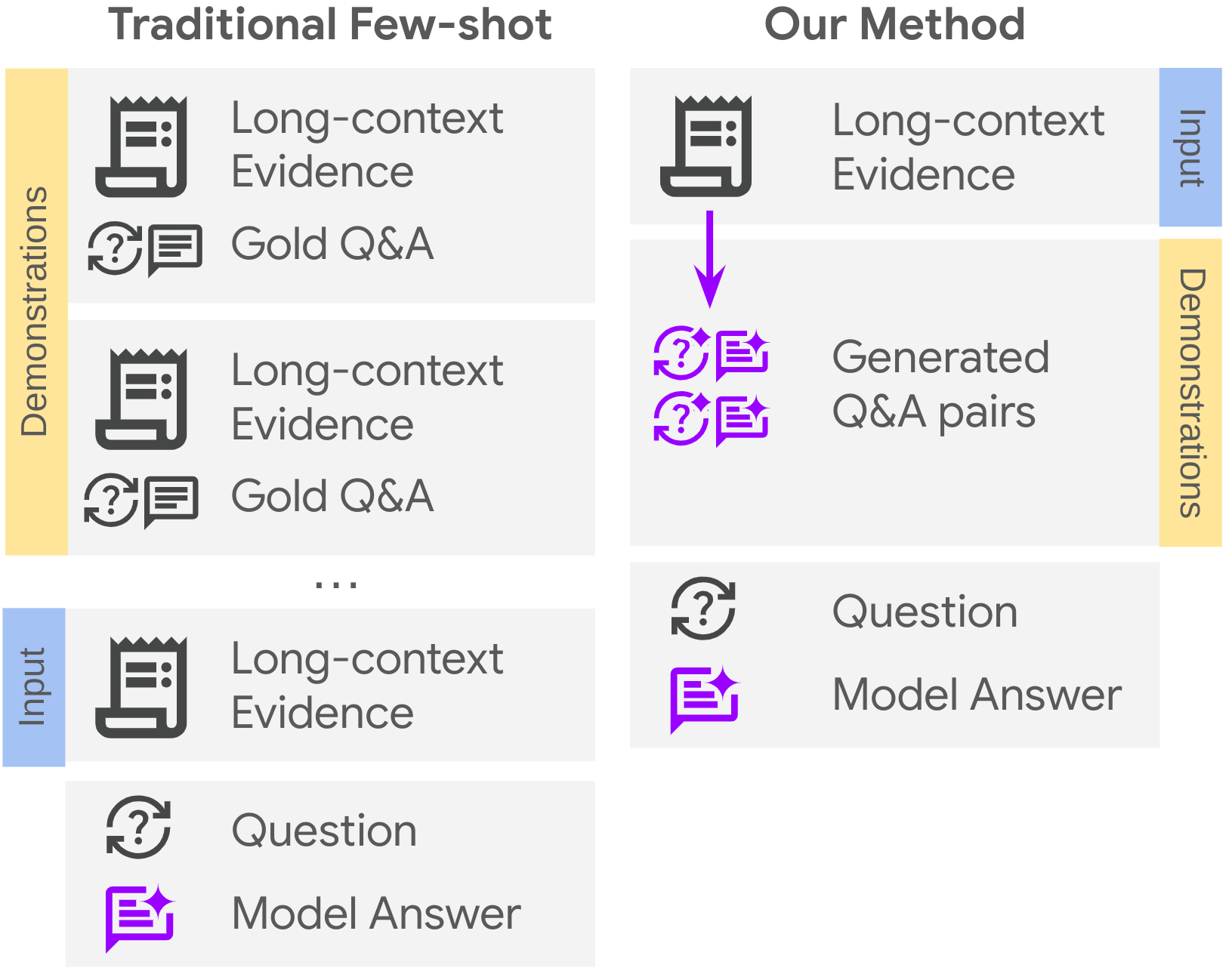}
    \caption{
    Comparison of traditional In-Context-Learning (ICL) and our new method. In traditional ICL (left), each example comprises a possibly lengthy context, accompanied by a query and an answer, typically derived from the training dataset. Conversely, our approach (right) simplifies each example to just a question and an answer, both of which are generated directly from the provided input context.
    }
    \label{fig:comparison}
\end{figure}

Yet, while Large Language Models (LLMs) have demonstrated impressive capabilities in a variety of tasks including answering questions requiring one or multiple reasoning steps, they often struggle to answer simple questions when faced with long contexts. 
Despite substantial engineering efforts~\citep{Chen2023LongLoRAEF} to extend the context window of LLMs to extremely long inputs (32k and even 1M tokens), these models continue to struggle with much shorter inputs, comprising only a few thousand tokens, especially when the relevant information is buried in the middle of the context~\citep{Liu2023LostIT} or obscured by numerous irrelevant details~\citep{Levy2024SameTM}.

In-Context Learning (ICL) with few-shot examples may be an appealing solution to enhance LLM performance in long contexts. However, applying ICL in real-world scenarios without access to training data introduces significant challenges. Developers need to maintain a demonstration pool for retrieving semantically similar demonstrations to any given query~\citep{liu-etal-2022-makes, rubin-etal-2022-learning}. Furthermore, ICL adds a substantial token overhead to the prompt, an issue that becomes even more pronounced with long-context inputs.

In this work, we introduce a novel method to enhance the QA performance of LLMs in long input setups. Our approach, termed \method{}, leverages LLMs' In-Context Learning capability and is based on two principles. 
First, instead of typical ICL, where each few-shot example is standalone with a separate lengthy context and a question-answer (QA) pair, we propose to \textit{recycle} the given input context and automatically generate few-shot examples from this context. Specifically, we randomly select a few paragraphs from the given input context and generate QA pairs for each passage. These generated QAs serve as demonstration examples and are placed between the input context and the target input question. Figure~\ref{fig:comparison} illustrates the differences between the traditional ICL with few-shot examples and \method{}.  
Second, we enhance each ICL demonstration by \textit{explicitly} instructing the model to identify relevant information prior to generating an answer. 
Explicitly identifying relevant passages can be regarded as a structured Chain of Thought that incentivizes the model to pinpoint relevant information before reasoning, an essential capability for long-context processing.

By generating few-shot demonstrations from various sections of the input context while instructing the model to identify relevant passages, \method{} encourages the model to develop deeper reading comprehension skills specific to the given input evidence. This, in turn, allows the model to answer subsequent queries with higher accuracy. 
\method{} presents several advantages.
First, recycling the same context for ICL demonstrations ensures that the few-shot examples refer to the exact same domain as the input question, thus obviating the need for external retrieval of similar demonstrations.
Also, in terms of efficiency, since each example does not include its own input context, our method adds to the original prompt a minimal number of tokens, resulting in a substantially cheaper inference than traditional ICL.
Finally, \method{} generates answers with attribution to relevant paragraphs, improving the model's lookup ability and offering transparency, which substantially simplifies human evaluation~\citep{Menick2022TeachingLM, gao-etal-2023-enabling, liu-etal-2023-evaluating, slobodkin-etal-2024-attribute}.

We applied \method{} to 12 LLMs, both commercial (Gemini Pro, Nano, Flash~\citep{Reid2024Gemini1U}; GPT-4~\citep{Achiam2023GPT4TR}) and open-source ranging from 2B to 70B parameters (Llama 3.1~\citep{Dubey2024TheL3}; Mistral~\citep{Jiang2023Mistral7}; Mixtral~\citep{Jiang2024MixtralOE}; Gemma,~\citep{Riviere2024Gemma2I}). We evaluate our method on 7 QA datasets with long inputs, including common multi-hop QA datasets. Our experiments demonstrate that with only 3 self-generated few-shot examples, \method{} consistently outperforms the baseline on our evaluation set by 16 absolute points on average across models. In addition, for some models, \method{} enhances the robustness to the position of the relevant information within the text.
Interestingly, while our few-shot examples focus on single-paragraph answers, \method{} generalizes well to multi-hop QAs and where the answer requires information from multiple passages.

%% file: sections/02_bg.tex
\section{Background}

\paragraph{Challenges in Long Context for Language Modeling.}

LLMs have been well-documented to struggle when input length grows~\citep{an2023leval}, and especially when it exceeds input lengths seen during training~\citep{Anil2022ExploringLG}. Various methods have been proposed to advance long-context capabilities: Architectural, e.g., to augment the embedding layer to cleverly extrapolate to unseen lengths~\citep{Vaswani2017AttentionIA,Press2021TrainST,caciularu-etal-2022-long, tan-etal-2024-lloco};  via data, e.g., to incorporate longer inputs and more challenging long-context scenarios into training~\citep{chen2024longlora, he-etal-2024-never, chen-etal-2024-long}; via attention intervention~\citep{hsieh-etal-2024-found} or by considering question likelihood as a signal for prompt reordering~\citep{Liu2024LikelihoodAA}. However, this challenging problem stubbornly remains in competitive models~\citep{Liu2023LostIT,bishop-etal-2024-longdocfactscore,Levy2024SameTM}. In contrast to the above methods, \method{} is a simple method that does not involve training or architectural changes.

Many benchmarks targetting long-context have been proposed, such as Scrolls and Zero-Scrolls~\citep{shaham-etal-2022-scrolls,shaham-etal-2023-zeroscrolls}, Loogle~\citep{Li2023LooGLECL}, LongBench~\citep{Bai2023LongBenchAB}, L-Eval~\citep{an-etal-2024-l}, inter alia. The problem of designing informative and reliable benchmarks in long-context is an an active, ever-changing area of research~\citep{Goldman2024IsIR,Yen2024HELMETHT}. We describe the most relevant evaluation benchmarks used in this work in Section~\ref{sec:experiments}.

\paragraph{In-Context Learning}

In-context Learning (ICL) consists of adding demonstrations to the prompt in order to steer or improve model behavior~\citep{min-etal-2022-rethinking}. These demonstrations are either hand-crafted~\citep{Song2022ACS}, or retrieved from a large set of training examples~\citep{liu-etal-2022-makes, rubin-etal-2022-learning, Paranjape2023ARTAM}.
While ICL provides a flexible approach to learn new tasks without updating parameters~\citep{Brown2020LanguageMA,Luo2024IncontextLW}, applying ICL in real-world scenarios is challenging notably because there is no available training data for each user query.

More closely related to our work, a few recent studies propose to prompt LLMs for automatically generating in-context demonstrations for various short context tasks. For instance, \citealp{Kim2022SelfGeneratedIL} focus on sentence classification tasks and prompt LLMs to generate full demonstrations conditioned on a label (e.g., ``write a negative review'') and  \citep{chen-etal-2023-self, yasunaga2024large, li2024are} generate relevant exemplars to the query for reasoning problems. While effective for reasoning tasks with short contexts, these methods are not directly applicable to long-context scenarios because LLMs would need to generate not just a question and an answer, but also an entire long context for each demonstration, which is both computationally expensive and prone to hallucination. 
\method{} addresses these challenges by introducing a novel and efficient strategy: rather than generating entirely new contexts, it \textit{recycles} the input context and generates only the question-answer (QA) pairs needed as demonstrations.

%% file: sections/03_method.tex
\section{\method{}}
\label{sec:method}

\begin{figure*}[t]
    \centering
    \includegraphics[width=.98\textwidth]{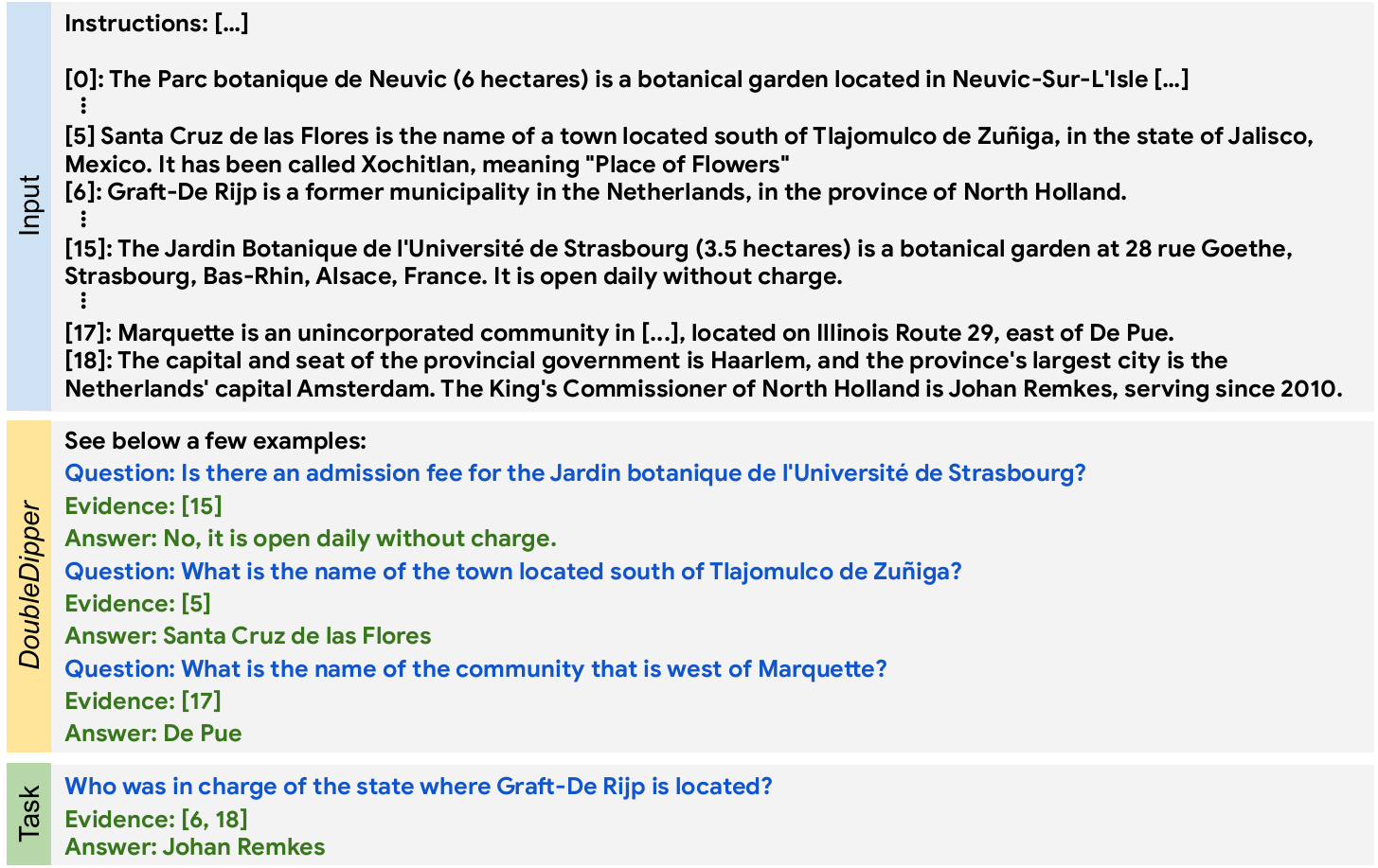}
    \caption{Example of \method{} applied to the MuSique dataset. Given 20 passages as input, \method{} randomly selects 3 passages (specifically passages 15, 5, 17) and automatically generates a question-pair for each one. As each QA is associated with its respective paragraph, we form the demonstrations to instruct the model to identify the relevant passage(s) and the correct answer. }
    \label{fig:doubledipper}
\end{figure*}

Given a long input text $\mathcal{C}$ composed of $n$ paragraphs $\mathcal{C} = \{p_1, p_2, ..., p_n\}$ and a question $q$, the goal is 
to generate the answer $a$ and identify the set(s) of paragraphs that support the answer $S = \{s_1, ..., s_k\}$. 
The number of the supporting paragraphs is not known in advance and can be one or more.

We describe \method{}, an efficient method for improving the performance of large language models (LLMs) when dealing with long contexts. The core principles of \method{} involve: (1) recycling the input context to automatically generate few-shot examples, and (2) ``teaching'' the model via in-context learning (ICL) to \textit{explicitly} pinpoint the supporting paragraphs before generating the answer.

Figure~\ref{fig:doubledipper} illustrates \method{}.
Starting with the input paragraphs $\mathcal{C}$, we initially select $k$ paragraphs at random (e.g., paragraphs $15$, $5$, and $17$, for $k:=3$). For each chosen paragraph, we prompt the model to formulate a question that pertains to the specific paragraph, accompanied by an appropriate answer (for further details on prompt specifications, refer to Appendix~\ref{app:prompts}). Each generated QA pair is directly associated with its origin paragraph, enabling us to assemble the following structured in-context demonstration, shown as the \method{} block in Figure~\ref{fig:doubledipper}:
\begin{align*}
\text{Question}&: q_i \\
\text{Evidence}&: p_i \\
\text{Answer}&: a_i
\end{align*}
Here, $p_i$ indicates the index of the paragraph associated with the QA pair $(q_i, a_i)$. 
Given a context $\mathcal{C}$ and a test question $q$, we compile a list of generated demonstrations $\mathcal{D}_\text{demo} = (q_1, p_1, a_1, \dots, q_k, p_k, a_k)$ to predict the output $y \sim p_\theta( y \mid \mathcal{C}, \mathcal{D}_\text{demo }, q)$ where the output $y$ is the concatenation of one or more indices of the supporting paragraph(s) $S$ and the answer $a$.

Unlike traditional few shot examples that instruct the model about a specific \textit{task}, \method{} aims to coach the model on how to ``handle'' the input context. This is achieved by guiding the model to explicitly localize relevant information before generating the answer. Also, by randomly sampling multiple paragraphs from the input, \method{} guarantees that the ICL demonstrations involve reading different parts of the context, allowing the model to better comprehend the input text. Beyond improving the performance of the QA task, instructing the model to provide the supporting paragraphs offers transparency and substantially eases human evaluation.

\method{} offers several advantages. First, as each example in the demonstration consists only of a question, an answer and the ID of relevant passage, the number of added tokens due to the extra demonstrations is minimal (5\%), leading to a low additional cost and computation compared to the traditional In-Context-Learning.
Furthermore, by reusing the same context to generate demonstrations, our approach guarantees that all few shot examples are derived from the exact same domain as the input query \citep{rubin-etal-2022-learning}.

%% file: sections/04_experiments.tex
\section{Experiments}

\label{sec:experiments}

\paragraph{Datasets}

\input{tables/datasets}

We apply our method to various datasets, each presenting its own domain-specific challenges. We selected these datasets because the supporting paragraphs are also annotated. Overall our evaluation set includes 5.5K instances, with statistics of each dataset given in Table~\ref{tab:data_stat}.

The Lost-in-the-middle dataset~\citep{Liu2023LostIT} includes examples from NaturalQuestions-Open~\citep{kwiatkowski-etal-2019-natural}. Each instance consists of twenty Wikipedia passages, with only one passage containing the answer to the query. The remaining passages are distractors that are lexically similar but do not contain the answer. To assess the robustness of LLMs to the position of relevant information, \citet{Liu2023LostIT} evaluated cases where the relevant passage appeared in positions 1, 5, 10, 15, and 20. Following their methodology, we sampled 500 instances for each position, resulting in a total of 2,500 instances.

FLenQA~\citep{Levy2024SameTM} is a benchmark that includes simple questions with answers of either ``True'' or ``False''  based on two key sentences. FLenQA includes three subtasks. The first subtask is MonoRel, where each instance asks whether a transitive relation between two entities holds based on the context (e.g., "Is X younger than Y?" based on the sentences "X is younger than Z" and "Z is younger than Y"). The second subtask, PIR, involves one key sentence indicating that a person is in a specific room and another key sentence describing a property of this room. The question asks whether the person is in a room with the described property. The final subtask is SRT, based on RuleTaker~\citep{Clark2020TransformersAS}. Each instance consists of a logical rule, two sentences each introducing a fact, and a question over the rule and facts. For each subtask, FLenQA includes contexts with varying lengths, from 50 to 3,000 tokens, by simply adding irrelevant text, demonstrating consistent performance degradation with increased input length. In our experiments, we sampled 250 instances for each subtask with input lengths of 2,000 and 3,000 tokens, leading to a total of 1,500 instances.

In addition, we evaluate our method on common multi-hop QA benchmarks. We sampled 500 instances from HotPotQA~\citep{yang-etal-2018-hotpotqa}, 2Wiki~\citep{ho-etal-2020-constructing}, and MuSiQue~\citep{Trivedi2021MM}. In all these datasets, the input text includes multiple passages, and models need to perform at least two steps of reasoning over different passages in order to answer the question.

\paragraph{Models}

We apply \method{} to a variety of models, both commercial and open-source. The commercial models include Gemini 1.5 Pro, Gemini 1.5 Flash~\citep{Reid2024Gemini1U} and GPT-4o-mini~\citep{Achiam2023GPT4TR}. The open-source models we tested are Llama 3.1 8B, Llama 3.1 70B~\citep{Dubey2024TheL3}, Gemma 2B (v2), Gemma 9B (v2) and Gemma 27B~\citep{Riviere2024Gemma2I}, Mistral-7B-Instruct (v0.2)~\citep{Jiang2023Mistral7}, Mixtral-8x7B-Instruct (v0.1)~\citep{Jiang2024MixtralOE} and Mistral Nemo Instruct 2407\footnote{\url{https://mistral.ai/news/mistral-nemo/}}.

Few-shot generation in \method{} is an auxiliary task and should ideally run in an efficient time without requiring heavy resources. Therefore, in our main experiments, we employ Gemma 2B to generate the demonstrations at it is the smallest and most efficient model used in our experiments. 
See Section~\ref{sec:analysis} for an ablation analysis of the effect of the chosen model for generating the demonstrations.

\paragraph{Baselines}

We evaluate \method{} against two main baselines. The first is a vanilla baseline that takes as input the entire context $\mathcal{C}$ and the query $q$ and generates only the answer $a$, a common prompting strategy in recent studies on long context~\citep{Liu2023LostIT, Levy2024SameTM}. The second baseline, \emph{Zero-shot + Evidence Retrieval}, prompts the model in a zero-shot manner to first identify relevant passages before generating the answer, following common practices in generating with attribution~\citep{gao-etal-2023-enabling, slobodkin-etal-2024-attribute, fierro-etal-2024-learning}.

\paragraph{Evaluation}

We evaluate each dataset with the original evaluation metrics. Namely, we report Accuracy for Lost-in-the-middle~\citep{Liu2023LostIT} and FLenQA~\citep{Levy2024SameTM}, and Token F1 for HotPotQA~\citep{yang-etal-2018-hotpotqa}, 2Wiki~\citep{ho-etal-2020-constructing} and MuSique~\citep{Trivedi2021MM}. 

In addition to the task's accuracy, we also evaluate the performance of the identification of the supporting paragraph(s), by computing the F1 score on the predicted set of supporting passages compared to the ground truth~\citep{yang-etal-2018-hotpotqa, ho-etal-2020-constructing, Trivedi2021MM}. 

\paragraph{Implementation Details}

We randomly select three passages from the input, each containing at least two sentences, and ask the model to generate a single QA pair for each passage. See Section~\ref{sec:analysis} for an analysis of the number of self-generated demonstrations on the performance

%% file: tables/datasets.tex
\begin{table}
    \centering
    
    \resizebox{0.48\textwidth}{!}{
    \begin{tabular}{lccc}
    \toprule
         \textbf{Dataset} & \# \textbf{Instances} & \textbf{Avg. \# tokens}  \\
         \midrule
         Lost-in-the-middle & 2,500 & 2,815 \\ 
         FLenQA & 1,500 & 3,225\\
         HotpotQA & 500 & 1,646 \\
         2Wiki & 500 & 1,222\\
         MuSiQue & 500 & 2,549 \\
         \bottomrule
    \end{tabular}}
    \caption{Evaluation datasets in our experiments. The average number of tokens is computed according to Gemma's tokenization of the vanilla prompt.}
    \label{tab:data_stat}
\end{table}

%% file: sections/05_results.tex
\section{Results}
\label{sec:results}

\input{tables/main_results_dd_with_gemma2}

\paragraph{Result 1: \method{} offers a substantial performance boost.}
Table~\ref{tab:main_results} presents the QA performance of the baseline, \emph{Zero-shot + Evidence Retrieval} and \method{} on our evaluation set. For brevity, we report the results of eight models here and four models in Appendix~\ref{app:additional_results} in Table~\ref{tab:appendix_results}, which show similar trends.
The results first show that prompting models to explicitly identify the relevant paragraphs before generating the answer (\emph{Zero-shot + Evidence Retrieval}) leads to a performance improvement of 9.7 points on average across models over the vanilla baseline.
\method{}, leveraging demonstrations generated with the efficient Gemma 2B parameters, offers an additional substantial boost of 6.3 points for all models on average, \textbf{culminating in an overall improvement of 16 absolute points over the vanilla baseline}. 
Notably, while \method{} produces simple QAs answerable from a single paragraph, \textit{it always surpasses the baseline in multi-hop QA datasets} (HotPotQA, 2Wiki and MuSique).
Likewise, \method{} outperforms the baseline also on most FLenQA datasets (PIR, MonoRel and SRT), which involve synthetic True/False questions although the demonstrations in \method{} are typically simple factoid questions. 

While \method{} exhibits strong performance overall, we observe nuanced behavior on the SRT dataset, with performance gains varying across models (improvement for Gemini, Llama 3.1 8B, Mistral 7B and Mistral Nemo). This discrepancy is likely due to the fact that SRT demands a specific type of reasoning, where models must reason over both a rule (e.g., ``If X is big and X is good then X is tall'') and dispersed facts (e.g., ``Erin is Good'' and ``Erin is furry'') to determine whether a statement (e.g., ``Erin is tall'') can be derived from the context. 
Finally, the Lost dataset highlights a specific characteristic of \method{}: while the baseline's known positional bias~\citep{liu-etal-2024-lost} masks the \textit{average} improvement for some models, \method{} substantially boost performance when relevant information appear in the \textit{middle} of the context (further elaborated in Result 3), demonstrating its efficacy in mitigating positional biases in long-context settings.

\input{tables/retrieval_f1}

\paragraph{Result 2: Learning to retrieve the evidence(s) with \method{} is more effective in commercial and large open source models.} 
Table~\ref{tab:retrieval} presents the performance of the supporting paragraphs prediction for the \emph{Zero-shot + Evidence Retrieval} and \method{} on our evaluation set. For all commercial models, Llama 3.1 70B and the recent Mistral-Nemo-Instruct-2407, \method{} predicts better the supporting paragraphs than in the zero-shot setting (+2.6 F1 for Gemini Pro, +3.4 F1 for Gemini Flash, +2.7 F1 for Llama 3.1 70B and +6.4 F1 for Mistral-Nemo-Instruct-2407). Conversely, \method{} slightly hurts the performance of common open source models (e.g., -2.8 F1 for Mistral, -0.3 F1 for Llama 3.1, -1.3 F1 for Gemma 27B, etc.).

This discrepancy appears to stem from shortcut learning \citep{tang-etal-2023-large}. Indeed, our demonstrations in \method{} use a single evidence paragraph, and smaller models tend to overfit to this pattern, learning to retrieve only one passage even when multiple are needed. For example, under \method{}, Gemma 2 9B retrieves only 1.2 paragraphs on average, compared to 2 in the ``unconstrained'' zero-shot setting. Larger models like Gemini Pro do not exhibit this behavior, correctly generalizing to predict an average of 2 evidence paragraphs even with single-paragraph demonstrations.
This hypothesis is strongly supported by the ``Lost'' dataset, which requires only a single evidence paragraph. On this dataset, nearly all models, including smaller ones, benefit substantially from \method{} (e.g., +12.2 F1 for Mistral 7B). 
Crucially, despite their suboptimal retrieval performance on multi-evidence tasks, these smaller models still produce better final answers with \method{} than the zero-shot baseline, highlighting the net benefit of the approach.

\paragraph{Result 3: \method{} makes models more robust to the position of relevant information.}

\begin{figure*}[t]
    \centering
    \includegraphics[width=.98\textwidth]{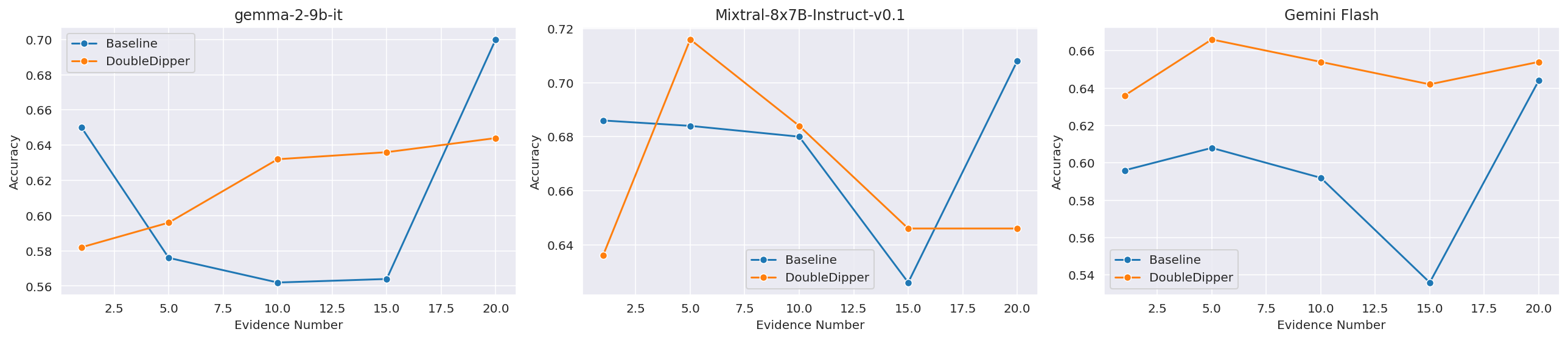}
    \caption{Performance (accuracy) of Gemma 2 9B, Mixtral 8x7B and Gemini Flash with and without \method{} on our sample of the Lost-in-the-middle dataset~\citep{Liu2023LostIT} according to the position of the document that contains the answer.}
    \label{fig:lost}
\end{figure*}

Following~\citet{Liu2023LostIT}, Figure~\ref{fig:lost} shows the performance of Gemma 2 9B, Mixtral 8x7B and Gemini Flash for both the baseline and \method{} on our sample of the Lost-of-the-middle dataset, according to the position of the document that contains the answer. See Appendix~\ref{app:lost} for the performance curve of the other tested models, which show similar trends. 

Overall, the performance curve for \method{} consistently surpasses the baseline when the relevant information appears ``in the middle'' and sometimes also at the beginning and/or the end (e.g., Gemini Flash). This variation can likely be attributed to the inherent biases of LLMs towards the beginning and end of inputs, while adding in context demonstrations mitigates this bias. This reveals that beyond improving performance, \method{} can make the model more robust to the position of the relevant document.

%% file: tables/main_results_dd_with_gemma2.tex
\begin{table*}[t!t]
\centering
\scriptsize

\resizebox{0.9\textwidth}{!}{
\begin{tabular}{lc|ccccccccc}\toprule
&\textbf{Avg.} &\textbf{2Wiki} &\textbf{MonoRel} &\textbf{PIR} &\textbf{SRT} &\textbf{HotPotQA} &\textbf{Lost} &\textbf{MuSique} \\\midrule
Gemini Pro (vanilla) &60.5 &24.9 &95.0 &97.6 &64.4 &46.7 &71.6 &23.1 \\
\quad \emph{Zero-shot + Evidence Retrieval} &62.3 &32.5 &94.6 &95.8 &62.4 &49.6 &\textbf{74.8} &26.5 \\
\quad \method{} &\textbf{70.4} &\textbf{46.8} &\textbf{97.4} &\textbf{99.0} &\textbf{79.6} &\textbf{60.9} & 72.4 &\textbf{36.4} \\
\midrule
Gemini Flash (vanilla)&42.9 &10.2 &70.0 &86.0 &57.6 &10.0 &59.5 &7.3 \\
\quad \emph{Zero-shot + Evidence Retrieval} &58.2 &30.2 &78.8 &90.6 &65.0 &44.9 &\textbf{67.4} &30.2 \\
\quad \method{} &\textbf{66.1} &\textbf{48.0} &\textbf{85.8} &\textbf{95.0} &\textbf{68.6} &\textbf{60.6} &65.0 &\textbf{39.7} \\
\midrule
Gemma 2 9B (v2) (vanilla)&44.0 &11.4 &74.8 &81.8 & 55.6 &13.8 &61.0 &9.3 \\
\quad \emph{Zero-shot + Evidence Retrieval} &58.7 &38.6 &82.0 &83.4 &\textbf{59.4} &56.1 &\textbf{64.4} &26.8 \\
\quad \method{} &\textbf{61.2} &\textbf{41.7} &\textbf{84.0} &\textbf{95.0} &51.4 &\textbf{61.2} & 61.8 &\textbf{33.3} \\
\midrule
Gemma 2 27B (v2) (vanilla) & 48.0 &11.2 &85.2 &79.2 & \textbf{63.8} &14.3 &62.4 &19.9 \\
\quad \emph{Zero-shot + Evidence Retrieval} &59.7 &34.3 &90.6 &87.4 &59.0 &51.7 &\textbf{65.2} &29.8 \\
\quad \method{} &\textbf{64.2} &\textbf{42.0} &\textbf{92.0} &\textbf{96.4} &58.6 &\textbf{64.2} & 62.5 &\textbf{33.9} \\
\midrule

Llama 3.1 8B (vanilla)&37.2 &11.8 &56.2 &52.2 &48.6 &20.7 &\textbf{63.5} &7.4 \\
\quad \emph{Zero-shot + Evidence Retrieval} &53.1 &\textbf{42.2} &71.8 &65.4 &50.8 &55.3 &62.0 &24.5 \\
\quad \method{} &\textbf{59.9} & 38.7 &\textbf{91.2} &\textbf{90.6} &\textbf{51.0} &\textbf{59.3} &58.1 &\textbf{30.1} \\
\midrule

Llama 3.1 70B (vanilla) & 67.5 &46.5 &93.2 &95.6 &\textbf{83.2} &57.0 &69.6 &27.6 \\
\quad \emph{Zero-shot + Evidence Retrieval} &71.0 &57.6 &97.8 &97.6 &81.0 &62.0 &\textbf{71.5} &29.5 \\
\quad \method{} &\textbf{72.9} &\textbf{62.3} &\textbf{98.6} &\textbf{98.8} &73.0 &\textbf{71.5} &66.0 &\textbf{40.2} \\
\midrule

Mistral 7B (v0.3) (vanilla)&37.4 &14.1 &59.2 &57.4 &50.2 &15.8 &\textbf{60.8} &4.6 \\
\quad \emph{Zero-shot + Evidence Retrieval} &44.0 &23.8 &66.2 &62.4 &49.6 &34.1 &58.6 &13.6 \\
\quad \method{} &\textbf{51.0} &\textbf{28.6} &\textbf{68.4} &\textbf{88.8} &\textbf{50.6} &\textbf{43.4} &60.7 &\textbf{16.7} \\
\midrule

Mistral-Nemo & 44.0 &17.2 &\textbf{72.6} &67.0 &51.0 &28.9 &59.7 &11.9 \\
\quad \emph{Zero-shot + Evidence Retrieval} &46.7 &29.1 &59.8 &67.2 &51.0 &39.8 &60.6 &19.4 \\
\quad \method{} &\textbf{53.3 }&\textbf{38.7} &58.0 &\textbf{81.0} &\textbf{51.4} &\textbf{51.9} &\textbf{62.9} &\textbf{29.5} \\
\bottomrule
\end{tabular}}
\caption{Accuracy of the QA task for the vanilla baseline (prompting the model to only answer the question), \emph{Zero-shot + Evidence Retrieval} (prompting the model to explicitly identify the relevant passage(s) before generating the answer) and \method{} with 3 demonstrations generated by Gemma 2 2B. }
\label{tab:main_results}
\end{table*}

%% file: tables/retrieval_f1.tex
\begin{table*}[t]\centering
\resizebox{0.95\textwidth}{!}{
\begin{tabular}{llccccccccc}\toprule
 & &\textbf{Avg.} &\textbf{2Wiki} &\textbf{MonoRel} &\textbf{PIR} &\textbf{SRT} &\textbf{HotPotQA} &\textbf{Lost} &\textbf{MuSique} \\\midrule
 \multirow{2}{*}{Gemini Pro} &\emph{Zero-shot + Evidence Retrieval} &83.7 &\textbf{96.7} &97.7 &\textbf{97.5} &62.8 &\textbf{92.1} &63.6 &75.3 \\
&\method{} &\textbf{86.3} &94.4 &\textbf{99.8} &97.1 &\textbf{80.9} &90.0 &\textbf{66.4} &\textbf{75.4} \\
\midrule
\multirow{2}{*}{Gemini Flash} &\emph{Zero-shot + Evidence Retrieval} &75.7 &82.3 &90.5 &72.6 &70.4 &80.9 &\textbf{67.3} &65.9 \\
&\method{} &\textbf{79.1} &\textbf{83.7} &\textbf{98.3} &\textbf{80.2} &\textbf{71.2} &\textbf{84.6} &66.1 &\textbf{69.5} \\
\midrule

\multirow{2}{*}{Gemma 2 9B} & \emph{Zero-shot + Evidence Retrieval} &\textbf{61.9} &\textbf{76.7} &\textbf{69.3} &\textbf{60.3} &\textbf{43.2} &76.5 &51.5 &55.9 \\
&\method{} &57.0 &74.2 &52.5 &59.0 &23.1 &\textbf{78.4} &\textbf{55.3} &\textbf{56.8} \\
\midrule
\multirow{2}{*}{Gemma 2 27B} & \emph{Zero-shot + Evidence Retrieval} &\textbf{85.1} &96.2 &\textbf{97.9} &96.1 &\textbf{84.1} &90.7 &53.4 &77.3 \\
&\method{} &83.8 &\textbf{97.5} &85.4 &\textbf{97.6} &74.9 &\textbf{93.0} &\textbf{59.7} &\textbf{78.8} \\
\midrule
\multirow{2}{*}{Llama 3.1 8B} &\emph{Zero-shot + Evidence Retrieval} &\textbf{61.7} &53.2 &\textbf{86.5} &\textbf{66.9} &\textbf{67.8} &63.2 &41.1 &53.5 \\
&\method{} &61.4 &\textbf{68.9} &71.4 &54.9 &52.8 &\textbf{73.7} &\textbf{53.0} &\textbf{54.8} \\
\midrule
\multirow{2}{*}{Llama 3.1 70B} &\emph{Zero-shot + Evidence Retrieval} & 85.1	&98.2&	98&	89.2&	82.3&	93.9	&52.5&	\textbf{81.6} \\
&\method{} & \textbf{87.8}&	\textbf{98.6}&	\textbf{100}	&\textbf{92.8}&	\textbf{83.8}&	\textbf{96.5}	&\textbf{61.6}&	\textbf{81.6} \\
\midrule
\multirow{2}{*}{Mistral 7B (v0.3)} &\emph{Zero-shot + Evidence Retrieval} &\textbf{46.6} &62.4 &\textbf{49.8} &\textbf{46.2} &\textbf{17.5} &64.0 &43.4 &\textbf{43.0} \\
&\method{} &43.8 &\textbf{63.4} &33.8 &42.5 &4.2 &\textbf{66.7} &\textbf{55.6} &40.2 \\
\midrule
\multirow{2}{*}{Mixtral 7x8B v(0.1)} &\emph{Zero-shot + Evidence Retrieval} &\textbf{60.0} &\textbf{72.4} &69.4 &\textbf{64.6} &\textbf{43.4} &\textbf{76.9} &40.6 &52.4 \\
&\method{} &58.9 &70.4 &\textbf{81.6} &63.6 &18.3 &75.0 &\textbf{50.4} &\textbf{53.2} \\
\midrule

\multirow{2}{*}{Mistral-Nemo} &\emph{Zero-shot + Evidence Retrieval} & 69.7	&91.1&	85.0	&76.9	&35.9	&88.8&	39.8&	70.3\\
&\method{}&\textbf{76.1}&	\textbf{95.9}&	\textbf{95.1}&\textbf{	81.5}&	\textbf{39.0}&	\textbf{93.7}&	\textbf{54.8}&\textbf{	73.0 }\\

\bottomrule
\end{tabular}}
\caption{Performance (F1) of supporting paragraph(s) prediction. }
\label{tab:retrieval}
\end{table*}

%% file: sections/06_analysis.tex
\section{Ablation Studies}
\label{sec:analysis}

\input{tables/few_shots}

\paragraph{How many examples are needed?}
In Table~\ref{tab:few_shots}, we explore for some models the impact of varying $k$, the number of self-generated few-shot examples in \method{} to 1, 3, 5, and 10. On average, a single demonstration already provides an improvement over the baseline. Three demonstrations adds another boost of 2 points, while increasing the number of demonstrations to 5 and 10 leads to a marginal improvement. 
This finding is in line with previous work~\citep{NEURIPS2020_1457c0d6,min-etal-2022-rethinking}. We conclude that a small number of examples carries most of the benefit with our method, but given additional computation budget, adding more examples does carry additional minor benefit.

\paragraph{\method{} without identification of supporting paragraphs}
To ablate the second core principle of \method{}—the explicit identification of supporting paragraphs prior to answer generation—we prompt open-source models with self-generated few-shot examples consisting solely of question-answer pairs, without instructing the model to retrieve the relevant passage(s). These demonstrations may undermine \method{}'s objective by encouraging models to produce answers without grounding them in the source text. The results confirms our hypothesis: removing evidence identification consistently degrades QA performance compared to the full \method{} approach. When averaging results across models and datasets, this omission leads to a substantial performance drop from 54.8 to 46.6. Detailed results are provided in Appendix~\ref{app:impact_retrieval}, Table~\ref{tab:dd_without_evidence}.

\input{tables/dd_few_shot_generator}
\paragraph{Investigating the effect of the few-shot generator}
To understand the impact of the default chosen model (Gemma 2 2B) for generating the demonstrations, we conducted two additional experiments. The first experiment is \textsc{Self} in which we use the same model for generating the demonstrations and for answering the original question. In the second experiment, we generate the demonstrations with the best LLM used in our experiments, namely Gemini Pro. The average results are reported in Table~\ref{tab:few_shot_generator} and the performance for each evaluation dataset is presented in Appendix~\ref{app:subsec:few_shot_model}. The results show that generating the demonstrations with Gemma 2 or \textsc{Self} achieves similar performance, while Gemini Pro leads to a consistent increase in performance across models, indicating that future better models can improve further the performance. Please refer to Appendix~\ref{app:subsec:few_shot_model} for additional ablations on the few-shot generation.

\input{tables/icl_vs_ddp}

\paragraph{\method{} vs. Traditional ICL} 
Another alternative to use ICL in practice is to preprend each QA prompt by a fixed set of QA demonstrations, each composed of a context, a question and an answer. Although the demonstrations are not necessarily from the same distribution as each user query, this common practice is helpful for task recognition and for an overview of the overall format~\citep{min-etal-2022-rethinking, pan-etal-2023-context}.
For the demonstrations, we randomly selected 3 examples from the SQuAD 2.0 dataset~\citep{rajpurkar-etal-2018-know}.
We compare the average results of the baseline, ICL and \method{} in Table~\ref{tab:icl_vs_dd_avg_results} and report full results in Appendix~\ref{app:icl_with_squad} in Table~\ref{tab:appendix_icl}. 
While ICL is effective and outperforms the baseline, \method{} provides an additional performance boost of 9.5 points on average.

\paragraph{Qualitative analysis: Correctness of the generated QA pairs }

We manually analyze 150 QAs generated by Gemma 2B as demonstrations. Our review confirms that 93.5\% of these self-generated QAs are correct, meaning that the question is meaningful and the answer could be found in the corresponding paragraph.

%% file: tables/few_shots.tex
\begin{table}
\centering

\resizebox{0.48\textwidth}{!}{
\begin{tabular}{lrrrrr}\toprule
&$k=1$ &$k=3$ &$k=5$ & $k=10$ \\\midrule
Gemini Nano &60.0 &62.1 &62.2 &62.3 \\
Gemini Flash  &65.3 &66.1 &65.9 &66.1 \\
Gemma 2B (v2)&47.0 &49.5 &49.6 &49.9 \\
Gemma 9B (v2) &58.7 &61.2 &61.4 &61.3 \\
Llama 3.1 &57.7 &59.9 &60.6 &61.4 \\
Mistral 7B (v0.3) &48.9 &51.0 &51.1 &51.4 \\
Mixtral 7x8B (v0.1) &49.3 &52.2 &51.7 &52.2 \\
\bottomrule
\end{tabular}}
\caption{ Average performance on our evaluation set with various numbers of self-generated few shot demonstrations ($k$) in \method{}. See Appendix~\ref{app:subsec:k} for the results on each evaluation dataset.}
\label{tab:few_shots}
\end{table}

%% file: tables/dd_few_shot_generator.tex
\begin{table}
\centering
\resizebox{0.42\textwidth}{!}{
\begin{tabular}{lccc}\toprule
&\textbf{Gemma 2B} &\textbf{Self} &\textbf{Gemini Pro} \\\midrule
Gemini Pro &70.4 &\textbf{71.6} &\textbf{71.6} \\
Gemini Flash &66.1 &67.5 &\textbf{68.1} \\
Gemini Nano &62.1 &61.7 &\textbf{62.8} \\
Gemma 2B &49.5 &49.5 &\textbf{51.0} \\
Gemma 9B &61.2 &62.0 &\textbf{63.5} \\
Llama 3.1 &59.9 &60.1 &\textbf{61.5} \\
Mistral v0.3 &51.0 &49.8 &\textbf{52.6} \\
Mixtral &52.2 &49.8 &\textbf{54.4} \\
\bottomrule
\end{tabular}}
\caption{Average performance of \method{} with different models for generating the demonstrations. See Appendix~\ref{app:subsec:few_shot_model} for the results on each evaluation dataset.}
\label{tab:few_shot_generator}
\end{table}

%% file: tables/icl_vs_ddp.tex
\begin{table}
\centering

\resizebox{0.47\textwidth}{!}{
\begin{tabular}{lccc}\toprule
& Baseline &ICL & \method{} \\\midrule
Gemma 2 9B & 44.0 &51.0 &\textbf{61.2} \\
Gemma 2 27B & 48.0 &54.9 &\textbf{64.2} \\
Llama 3.1 8B & 37.2 &40.7 &\textbf{59.9} \\
Llama 3.1 70B & 67.5 &65.4 &\textbf{72.9} \\
Mistral v0.3 & 37.4 &42.0 &\textbf{51.0} \\
Mixtral v0.1 & 42.6 &45.1 &\textbf{52.2} \\
GPT 4o mini & 51.3 & 56.1 &\textbf{60.8} \\
\bottomrule
\end{tabular}}
\caption{Comparison of traditional In-Context Learning (ICL) where each demonstration example comprises a full text, a question and an answer from an external dataset to \method{} where the demonstrations contain only question-answer pairs, automatically generated on the same input text.}
\label{tab:icl_vs_dd_avg_results}
\end{table}

%% file: sections/07_conclusion.tex
\section{Conclusion}
\label{sec:conclusion}

We introduce \method{}, a simple method for enhancing the performance of LLMs with long context. By recycling the input context to generate the demonstrations, \method{} successfully addresses the practical challenges of ICL with long-context and outperforms multiple baselines in various QA settings, including distractor passages in the input, True/False questions and multi-hop QA.

%% file: sections/08_limitations.tex
\section{Limitations}
\label{sec:limitations}

Our work has several limitations. 

First, our current work focuses on multiple variants of QA tasks (distractor passages, True/False and multi-hop), where the demonstrations teach the models to identify specific evidence paragraph(s) and extract answers. Future research can extend our work to other QA settings (e.g., information seeking) and additional tasks (e.g., summarization). 

Second, our evaluation set is constrained to instances that are solely in English and range between 1,000 to 4,000 tokens. While this demonstrates the method's effectiveness, its scalability and performance on much longer contexts (e.g., 100k+ tokens) and in multilingual settings remain open questions.

Finally, while \method{} is significantly more token-efficient than traditional ICL, the initial step of generating demonstrations introduces a computational and latency overhead compared to a zero-shot baseline. This presents a trade-off between inference cost and the substantial performance gains our method provides.

%% file: prompts/qg_prompt.tex
\begin{figure*}[t]

\lstdefinestyle{promptStyle}
{
    basicstyle={\footnotesize\ttfamily},%
    xleftmargin=2.8em,%
    xrightmargin=1.5em,
    showstringspaces=false,
      showspaces=false,
        showtabs=false,
    tabsize=2,
    breaklines=true,
        flexiblecolumns=true,
        escapeinside={<@}{@>},
          breakatwhitespace=true
}

\newtcblisting{mylisting}[1]{
  enhanced,
  listing only,
  boxrule=0.8pt,
  sharp corners=downhill,
  top=0mm,
  bottom=0mm,
  left=2mm,
  right=0mm,
  boxsep=0mm,
  colframe=black,
  colback=white,
  listing options={
    style=#1
  }
}

\definecolor{instructionsColor}{rgb}{0.1, 0.5, 0.1}

\begin{mylisting}{promptStyle}
<@\textcolor{instructionsColor}{
Given the following TEXT, please write a simple question whose answer appears verbatim in the text.
The question should include enough information so that it can be understood without the text. 
The answer should be concise.
Please write both the question ans answer in the following format:}@>
<@\color{blue}Q:@>
<@\color{blue}A:@> 
TEXT: [PARAGRAPH]
\end{mylisting}
\caption{Template prompt for generating the QA pairs in \method{}.}
\label{fig:qg_prompt}
\end{figure*}

%% file: tables/appendix_results_dd_vs_baseline.tex
\begin{table*}[t!t]
\centering
\scriptsize

\resizebox{0.9\textwidth}{!}{
\begin{tabular}{lc|ccccccccc}\toprule
&\textbf{Avg.} &\textbf{2Wiki} &\textbf{MonoRel} &\textbf{PIR} &\textbf{SRT} &\textbf{HotPotQA} &\textbf{Lost} &\textbf{MuSique} \\\midrule
Gemini Nano (vanilla)&41.6 &10.8 &72.2 &66.8 &55.4 &21.3 &59.6 &5.2 \\
\quad \emph{Zero-shot + Evidence Retrieval} & 56.5 &32.0 &82.4 &82.4 &\textbf{56.4} &56.7 &\textbf{60.5} &25.2 \\
\quad \method{} &\textbf{62.1} &\textbf{40.6} &\textbf{86.6} &\textbf{95.4} & 56.2 &\textbf{65.1} &\textbf{60.5} &\textbf{30.4} \\
\midrule

GPT-4o-mini (vanilla) & 51.3 &19.0 &78.6 &89.6 &\textbf{71.6} &23.3 &67.8 &9.3 \\
\quad \emph{Zero-shot + Evidence Retrieval} & 53.8 &19.4 &89.6 &93.0 &63.2 &26.2 &\textbf{67.9} &17.6 \\
\quad \method{} &\textbf{60.8} &\textbf{29.8} &\textbf{94.8} &\textbf{96.4} &61.6 &\textbf{53.7} &64.7 &\textbf{24.5} \\
\midrule
Gemma 2 2B (v2) (vanilla) &38.6 &8.9 &71.8 &68.6 &\textbf{51.2} &13.3 &\textbf{49.6} &6.5 \\
\quad \emph{Zero-shot + Evidence Retrieval} &42.0 &22.3 &66.8 &70.6 &40.2 &30.6 &47.6 &16.2 \\
\quad \method{} &\textbf{49.5} &\textbf{23.7} &\textbf{85.8} &\textbf{81.6} &50.0 &\textbf{39.9} &46.7 &\textbf{18.8} \\

\midrule
Mixtral 7x8B (v0.1) (vanilla) &42.6 &13.7 &73.0 &66.2 &\textbf{51.0} &18.2 & 67.7 &8.4 \\
\quad \emph{Zero-shot + Evidence Retrieval} &47.4 &18.8 &81.8 &73.6 &50.6 &26.3 &\textbf{67.9} &13.1 \\
\quad \method{} &\textbf{52.2} &\textbf{22.3} &\textbf{91.8} &\textbf{86.0} &47.8 &\textbf{35.1} &66.6 &\textbf{16.0}\\

\bottomrule
\end{tabular}}
\caption{Accuracy of the QA task for the vanilla baseline (prompting the model to only answer the question), \emph{Zero-shot + Evidence Retrieval} (prompting the model to explicitly identify the relevant passage(s) before generating the answer) and \method{} with 3 demonstrations generated by Gemma 2 2B. }
\label{tab:appendix_results}
\end{table*}

%% file: tables/few_shots_detailed.tex
\begin{table*}[!htp]\centering
\scriptsize

\begin{tabular}{llrrrrrrrrr}\toprule
& &Avg. &2Wiki &MonoRel &PIR &SRT &HotPotQA &Lost &MuSique \\\midrule
\multirow{4}{*}{Gemini Nano} &$k=1$ &60.03 &37.68 &85.20 &88.20 &56.40 &62.50 &60.68 &29.56 \\
&$k=3$ &62.12 &40.55 &86.60 &95.40 &56.20 &65.12 &60.52 &30.44 \\
&$k=5$ &62.25 &41.79 &87.00 &95.60 &55.20 &65.05 &60.56 &30.52 \\
&$k=10$ &62.33 &43.16 &86.00 &96.20 &55.20 &65.37 &60.44 &29.96 \\ 
\midrule
\multirow{4}{*}{Gemini Flash} &$k=1$ &65.34 &48.83 &84.80 &90.80 &66.40 &60.17 &65.16 &41.19 \\
&$k=3$ &66.11 &48.03 &85.80 &95.00 &68.60 &60.58 &65.04 &39.71 \\
&$k=5$ &65.87 &47.49 &87.20 &95.00 &66.60 &61.13 &64.20 &39.48 \\
& $k=10$ &66.08 &46.13 &86.20 &95.20 &69.00 &62.03 &63.80 &40.18 \\
\midrule
\multirow{4}{*}{Gemma 2B (v2)} &$k=1$ &47.05 &24.05 &73.60 &77.00 &50.20 &41.32 &47.04 &16.13 \\
&$k=3$ &49.48 &23.66 &85.80 &81.60 &50.00 &39.85 &46.68 &18.77 \\
&$k=5$ &49.59 &26.70 &85.40 &80.40 &48.40 &40.34 &46.56 &19.30 \\
&$k=10$ &49.92 &26.43 &85.80 &81.40 &49.00 &40.91 &47.68 &18.26 \\
\midrule
\multirow{4}{*}{Gemma 9B (v2)} &$k=1$ &58.77 &37.58 &81.00 &87.60 &51.60 &58.46 &63.44 &31.68 \\
&$k=3$ &61.20 &41.70 &84.00 &95.00 &51.40 &61.21 &61.80 &33.31 \\
&$k=5$ &61.40 &44.42 &84.20 &94.80 &50.00 &62.99 &60.76 &32.60 \\
&$k=10$ &61.34 &43.78 &84.40 &95.80 &50.40 &63.01 &59.96 &32.05 \\
\midrule
\multirow{4}{*}{Llama 3.1 8B} &$k=1$ &57.74 &38.20 &85.40 &84.80 &51.80 &57.50 &60.96 &25.49 \\
&$k=3$ &59.86 &38.75 &91.20 &90.60 &51.00 &59.29 &58.12 &30.09 \\
&$k=5$ &60.58 &41.56 &89.80 &91.40 &51.00 &61.31 &57.68 &31.28 \\
&$k=10$ &61.41 &44.68 &89.00 &91.00 &51.00 &63.88 &56.36 &33.98 \\
\midrule
\multirow{4}{*}{Mistral 7B} &$k=1$ &48.91 &28.86 &67.60 &77.40 &50.00 &40.62 &62.00 &15.91 \\
&$k=3$ &51.03 &28.65 &68.40 &88.80 &50.60 &43.44 &60.68 &16.66 \\
&$k=5$ &51.12 &29.86 &69.20 &89.00 &49.40 &43.83 &60.36 &16.23 \\
&$k=10$ &51.44 &30.14 &68.20 &89.40 &49.00 &46.49 &59.96 &16.91 \\
\midrule
\multirow{4}{*}{Mixtral 7x8B} &$k=1$ &49.26 &19.83 &86.00 &74.80 &49.40 &31.73 &69.00 &14.07 \\
&$k=3$ &52.22 &22.31 &91.80 &86.00 &47.80 &35.10 &66.56 &15.97 \\
&$k=5$ &51.65 &23.55 &91.40 &82.40 &47.20 &34.64 &64.92 &17.46 \\
&$k=10$ &52.18 &23.49 &91.80 &84.40 &47.00 &36.23 &64.00 &18.32 \\
\bottomrule
\end{tabular}
\caption{Performance of \method{} on our evaluation set with various numbers of self-generated few shot demonstrations ($k$).}
\label{tab:results_k_detailed}
\end{table*}

%% file: tables/dd_few_shot_generator_detailed.tex
\begin{table*}[!htp]\centering
\scriptsize
\begin{tabular}{llr|cccccccc}\toprule
& &Avg. &2Wiki &MonoRel &PIR &SRT &HotPotQA &Lost &MuSique \\\midrule
\multirow{2}{*}{Gemini Pro} &Gemma 2B &70.4 &46.8 &97.4 &99.0 &\textbf{79.6} &60.9 &72.4 &36.4 \\
&Self &\textbf{71.6} &\textbf{49.6} &\textbf{98.2} &\textbf{99.6} &78.8 &\textbf{62.5} &\textbf{72.5} &\textbf{40.1} \\
\midrule
\multirow{3}{*}{Gemini Flash} &Gemma 2B &66.1 &48.0 &85.8 &95.0 &68.6 &60.6 &65.0 &39.7 \\
&Self &67.5 &49.5 &88.6 &\textbf{96.8} &65.6 &64.3 &\textbf{65.2} &\textbf{42.5} \\
&Gemini Pro &\textbf{68.1} &\textbf{50.3} &\textbf{90.4} &96.0 &\textbf{69.0} &\textbf{64.8} &64.8 &41.6 \\
\midrule
\multirow{3}{*}{Gemini Nano} &Gemma 2B &62.1 &\textbf{40.6} &86.6 &95.4 &\textbf{56.2} &65.1 &60.5 &30.4 \\
&Self &61.7 &39.8 &85.8 &95.2 &\textbf{56.2} &64.3 &\textbf{60.7} &30.0 \\
&Gemini Pro &\textbf{62.8} &40.4 &\textbf{87.2} &\textbf{99.0} &54.4 &\textbf{66.8} &60.4 &\textbf{31.7} \\
\midrule
\multirow{2}{*}{Gemma 2 2B} &Self &49.5 &23.7 &\textbf{85.8} &81.6 &\textbf{50.0} &39.9 &\textbf{46.7} &\textbf{18.8} \\
&Gemini Pro &\textbf{51.0} &\textbf{25.2} &83.8 &\textbf{93.4} &48.0 &\textbf{41.7} &46.2 &18.4 \\
\midrule
\multirow{3}{*}{Gemma 2 9B} &Gemma 2B &61.2 &41.7 &84.0 &95.0 &51.4 &61.2 &61.8 &33.3 \\
&Self &62.0 &44.8 &82.4 &97.0 &50.6 &62.9 &61.5 &\textbf{34.9} \\
&Gemini Pro &\textbf{63.5} &\textbf{46.2} &\textbf{86.4} &\textbf{99.2} &\textbf{52.2} &\textbf{64.4} &\textbf{61.4} &34.8 \\
\midrule
\multirow{3}{*}{Llama 3.1 8B} &Gemma 2B &59.9 &38.7 &\textbf{91.2} &90.6 &51.0 &59.3 &\textbf{58.1} &30.1 \\
&Self &60.1 &\textbf{41.3} &88.6 &89.8 &50.8 &61.5 &57.8 &31.0 \\
&Gemini Pro &\textbf{61.5} &41.0 &89.8 &\textbf{95.6} &\textbf{51.6} &\textbf{63.5} &57.5 &\textbf{31.7} \\
\midrule
\multirow{3}{*}{Mistral 7B (v0.3)} &Gemma 2B &51.0 &28.6 &68.4 &88.8 &\textbf{50.6} &43.4 &60.7 &16.7 \\
&Self &49.8 &25.2 &\textbf{68.8} &93.8 &49.6 &36.5 &60.6 &14.3 \\
&Gemini Pro &\textbf{52.6} &\textbf{31.4} &67.2 &\textbf{96.4} &48.2 &\textbf{45.7} &\textbf{62.1} &\textbf{17.0} \\
\midrule
\multirow{3}{*}{Mixtral 7x8B (v0.1)} &Gemma 2B &52.2 &\textbf{22.3} &91.8 &86.0 &47.8 &35.1 &66.6 &16.0 \\
&Self &49.8 &16.0 &91.8 &86.0 &48.8 &25.5 &\textbf{67.8} &12.8 \\
&Gemini Pro &\textbf{54.4} &20.8 &\textbf{94.8} &\textbf{96.2} &\textbf{49.0} &\textbf{36.3} &66.6 &\textbf{16.9} \\
\bottomrule
\end{tabular}
\caption{Performance of \method{} according to the model for generating the demonstrations (Gemma 2B, Self or Gemini Pro).}
\label{tab:few_shot_generator_detailed}
\end{table*}

%% file: tables/dd_without_evidence.tex
\begin{table*}[!htp]\centering
\scriptsize
\begin{tabular}{llccccccccc}\toprule
& &Avg. &2Wiki &MonoRel &PIR &SRT &HotPotQA &Lost &MuSique \\\midrule
\multirow{2}{*}{Gemma 2B} &
\method{} (QA only) &46.4 &16.0 &84.6 &79.2 &49.6 &31.1 &\textbf{49.5} &14.7 \\
& \method{} &\textbf{49.5} &\textbf{23.7} &\textbf{85.8} &\textbf{81.6} &\textbf{50.0} &\textbf{39.9} &46.7 &\textbf{18.8} \\
\midrule
\multirow{2}{*}{Gemma 9B} & \method{} (QA only) &55.1 &22.5 &\textbf{91.0} &\textbf{96.6} &\textbf{52.0} &42.9 &61.0 &20.0 \\
&\method{} &\textbf{61.2} &\textbf{41.7} &84.0 &95.0 &51.4 &\textbf{61.2} &\textbf{61.8} &\textbf{33.3} \\
\midrule
\multirow{2}{*}{Mistral 7B (v0.3)} & \method{} (QA only) &49.1 &21.8 &\textbf{73.4} &\textbf{89.8} &48.6 &38.6 &60.3 &11.5 \\
&\method{} &\textbf{51.0} &\textbf{28.6} &68.4 &88.8 &\textbf{50.6} &\textbf{43.4} &\textbf{60.7} &\textbf{16.7} \\
\midrule
\multirow{2}{*}{Mixtral 8x7B (v0.1)} & \method{} (QA only) &49.5 &17.0 &\textbf{91.4} &\textbf{86.8} &50.8 &22.9 &65.8 &11.9 \\
&\method{} &\textbf{52.2} &\textbf{22.3} &91.8 &86.0 &\textbf{47.8} &\textbf{35.1} &\textbf{66.6} &\textbf{16.0} \\
\midrule
\multirow{2}{*}{Llama 3.1 8B} & \method{} (QA only) &32.7 &25.0 &9.2 &41.4 &46.6 &33.5 &56.7 &16.6 \\
& \method{} &\textbf{59.9} &\textbf{38.7} &\textbf{91.2} &\textbf{90.6} &\textbf{51.0} &\textbf{59.3} &\textbf{58.1} &\textbf{30.1} \\
\bottomrule
\end{tabular}
\caption{Performance of \method{} and \method{} without instructing the models to retrieve the evidence (QA only) on the QA datasets. }
\label{tab:dd_without_evidence}
\end{table*}

%% file: tables/appendix_icl_squad_results.tex
\begin{table*}[!t]
\centering
\scriptsize
\resizebox{0.8\textwidth}{!}{
\begin{tabular}{lccccccccc}\toprule
&Avg. &2Wiki &MonoRel &PIR &SRT &HotPotQA &Lost &MuSique \\\midrule
Gemma 2 2B &40.4 &11.5 &69.4 &59.8 &50.6 &33.1 &48.8 &9.7 \\
Gemma 2 9B &51.0 &21.7 &76.8 &78.8 &53.6 &49.1 &60.0 &17.3 \\
Gemma 2 27B &54.9 &21.3 &85.0 &76.6 &58.2 &53.3 &61.5 &28.3 \\
Llama 3.1 8B &40.7 &16.4 &59.6 &53.0 &45.6 &36.8 &62.5 &11.0 \\
Llama 3.1 70B &65.4 &38.1 &93.0 &91.4 &75.4 &59.7 &67.6 &32.4 \\
Mistral v0.3 &42.0 &25.5 &56.4 &52.8 &50.2 &42.7 &58.2 &8.3 \\
Mixtral v0.1 &45.1 &19.6 &73.6 &65.4 &50.4 &30.7 &66.6 &9.6 \\
GPT 4o mini &56.1 &29.7 &77.4 &86.8 &65.2 &49.4 &65.6 &18.5 \\
\bottomrule
\end{tabular}}
\caption{Accuracy of the QA task for the ICL experiment with 3 fixed demonstrations prepended to each instance. }
\label{tab:appendix_icl}
\end{table*}